\pdfoutput=1

\documentclass[11pt]{article}
\usepackage{ifpdf}
\usepackage[preprint]{coling}

\usepackage{times}
\usepackage{latexsym}

\usepackage[T1]{fontenc}

\usepackage[utf8]{inputenc}

\usepackage{microtype}

\usepackage{inconsolata}

\usepackage{graphicx}

\usepackage{booktabs}
\usepackage{listings}

\usepackage{array}
\usepackage{caption}
\usepackage{adjustbox}
\usepackage{enumerate}
\usepackage{enumitem}

\usepackage{comment}
\usepackage{tabularx}
\usepackage{amsmath}
\usepackage{rotating}
\usepackage{makecell}
\usepackage{float}
\usepackage{tablefootnote}

\lstdefinelanguage{json}{
    basicstyle=\ttfamily\small,
    showstringspaces=false,
    breaklines=true,
    frame=none,
    numbers=none,
    literate=
     *{0}{{0}}{1}
      {1}{{1}}{1}
      {2}{{2}}{1}
      {3}{{3}}{1}
      {4}{{4}}{1}
      {5}{{5}}{1}
      {6}{{6}}{1}
      {7}{{7}}{1}
      {8}{{8}}{1}
      {9}{{9}}{1}
      {:}{{:}}{1}
      {,}{{,}}{1}
      {\{}{{\{}}{1}
      {\}}{{\}}}{1}
      {[}{{[}}{1}
      {]}{{]}}{1},
}

\definecolor{pastelblue}{RGB}{99, 155, 255}
\definecolor{pastelpurple}{RGB}{178, 102, 255}
\definecolor{pastelgray}{RGB}{150, 150, 150}
\definecolor{pastelred}{RGB}{255, 102, 102}
\definecolor{pastelgreen}{RGB}{102, 255, 178}

\lstdefinestyle{custompython}{
  language=Python,
  basicstyle=\ttfamily\small,
  keywordstyle=\color{pink}, 
  stringstyle=\color{teal}, 
  commentstyle=\color{pastelgray}, 
  numberstyle=\tiny\color{pastelgray}, 
  identifierstyle=\color{black}, 
  keywords=[2]{_instruction_format}, 
  keywordstyle=[2]\color{purple},
  keywords=[3]{self}, 
  keywordstyle=[3]\color{olive},
  keywords=[4]{sys_message, query, assistant}, 
  keywordstyle=[4]\color{black},
  literate=
    *{0}{{{\color{pastelblue}0}}}{1}
     {1}{{{\color{pastelblue}1}}}{1}
     {2}{{{\color{pastelblue}2}}}{1}
     {3}{{{\color{pastelblue}3}}}{1}
     {4}{{{\color{pastelblue}4}}}{1}
     {5}{{{\color{pastelblue}5}}}{1}
     {6}{{{\color{pastelblue}6}}}{1}
     {7}{{{\color{pastelblue}7}}}{1}
     {8}{{{\color{pastelblue}8}}}{1}
     {9}{{{\color{pastelblue}9}}}{1}
     {:}{{:}}{1} 
     {,}{{,}}{1} 
     {\{}{{{\color{pastelgray}\{}}}{1}
     {\}}{{{\color{pastelgray}\}}}}{1}
     {[}{{{\color{pastelgray}[}}}{1}
     {]}{{{\color{pastelgray}]}}}{1},
  showstringspaces=false,
  breaklines=true,
  frame=none,
}

\title{Bilingual Evaluation of Language Models on General Knowledge in University Entrance Exams with Minimal Contamination}

\author{Eva Sánchez Salido,  Roser Morante, Julio Gonzalo, Guillermo Marco \\ {\bf Jorge Carrillo-de-Albornoz, Laura Plaza, Enrique Amigó, Andrés Fernández}   \\ {\bf Alejandro Benito-Santos, Adrián Ghajari Espinosa and Victor Fresno} \\
UNED Research Center in Natural Language Processing and Information Retrieval\thanks{\href{https://sites.google.com/view/nlp-uned/home}{nlp.uned.es}}\\
ETSI Informática, UNED - Juan del Rosal, 16 28040 Madrid, Spain\\
\small{\textbf{Correspondence:} \href{mailto:julio@lsi.uned.es}{julio@lsi.uned.es}}
}

\begin{document}
\pagestyle{empty} 

\maketitle
\begin{abstract}
In this article we present {\em UNED-ACCESS 2024}, a bilingual dataset that consists of 1003 multiple-choice questions of university entrance level exams in Spanish and English. Questions are originally formulated in Spanish and manually translated into English, and have not ever been publicly released, ensuring minimal contamination when evaluating Large Language Models with this dataset. A selection of current open-source and proprietary models are evaluated in a uniform zero-shot experimental setting both on the UNED-ACCESS 2024 dataset and on an equivalent subset of MMLU questions. Results show that 
(i) Smaller models not only perform worse than the largest models, but also degrade faster in Spanish than in English. The performance gap between both languages is negligible for the best models, but grows up to 37\% for smaller models; (ii) Model ranking on UNED-ACCESS 2024 is almost identical (0.98 Pearson correlation) to the one obtained with MMLU (a similar, but publicly available benchmark), suggesting that contamination affects similarly to all models, and (iii) As in publicly available datasets, reasoning questions in UNED-ACCESS are more challenging for models of all sizes. 
\end{abstract}

\section{Introduction}

With the recent progress in broadening the generalisation capabilities of Large Language Models (LLMs), much current research focuses on understanding their capabilities and limitations. Evaluation of generative models, such as Llama~\citep{touvron2023llama2openfoundation}, Mistral~\citep{jiang_mistral_2023}, Mixtral~\citep{jiang2024mixtralexperts}, Gemini~\citep{geminiteam2024geminifamilyhighlycapable}, Gemma~\citep{gemmateam2024gemmaopenmodelsbased}, GPT-3.5~\citep{brown2020languagemodelsfewshotlearners}, GPT-4 and GPT-4o ~\citep{openai2024gpt4technicalreport}, attempt at measuring their world knowledge, memorization and inference capabilities with benchmarks designed for these purposes. Although many benchmarks have been proposed as single-task evaluations,  with the emergence of general language models such as BERT~\citep{devlin-etal-2019-bert}, the development of more comprehensive benchmarks to measure the general capabilities of these models in a multi-task setting became popular (GLUE~\citep{wang-etal-2018-glue}, SuperGLUE~\citep{wang2020supergluestickierbenchmarkgeneralpurpose},  Big-Bench~\citep{srivastava_beyond_2022}, Big-Bench Hard~\citep{suzgun2022challengingbigbenchtaskschainofthought},  HELM~\citep{liang_holistic_2023}). 
A lot of emphasis is now put on assessments focusing on human-level cognitive tasks in real-world evaluation scenarios, such as exams. In this context, multiple-choice questions have emerged as one of the preferred methods for evaluating new generative models with datasets such as RACE~\citep{lai_race_2017}, MMLU~\citep{hendrycks_measuring_2021} or AGIEval~\citep{zhong_agieval_2023}, which includes university entrance exams, law school entrance exams, mathematics competitions and lawyer qualification tests. 
In this paper we present a new bilingual dataset of {\bf private} exam questions, UNED-ACCESS 2024. The dataset contains 1003 multiple-choice questions in Spanish from various subjects of the UNED Access Course for Over-25s in Spanish, together with high-quality English professional translations.
The dataset has been compiled at Universidad Nacional de Educación a Distancia (UNED), in the framework of the ODESIA project (\textit{Espacio de Observación de Inteligencia Artificial en Español}). This dataset is complementary to the ODESIA Benchmark, a collection of comparable datasets in English and Spanish to evaluate multiple NLP discriminative tasks. The ODESIA Benchmark is used in the ODESIA Leaderboard,\footnote{\url{https://leaderboard.odesia.uned.es}} with the aim of measuring the effectiveness gap between Spanish and English LLMs.
Two characteristics make this dataset unique: first, the contamination level is minimal, because UNED typically does not release the answers to the exam questions, which are only accessible to the teachers of each course. 
Second, this is a high-quality bilingual dataset, with  original questions in Spanish translated into English manually by a professional translators who did not use any external software. 

We also present results on evaluating multiple commercial and open-source LLMs on this dataset, and on an equivalent version of the popular MMLU dataset ~\citep{hendrycks_measuring_2021} with the same evaluation protocol. These allows us to address three research questions: \textbf{RQ1 (Contamination effects)}: How do current generative models perform on official private entrance university exams, in a minimal contamination setup? \textbf{RQ2 (Language effects)}: How do models perform comparatively in Spanish and English? and \textbf{RQ3 (Dataset size effect)}: How suitable is the zero-shot approach to evaluate LLMs on a relatively small dataset of exam questions?  

Our main contributions are: (i) The creation of a new, private benchmark with minimal contamination for the evaluation of generative LLMs on entrance university exams in Spanish (native) and English (translated);  (ii) A systematic evaluation of LLMs with the same prompts and hyperparameters for all models, in English and Spanish; (iii) Empirical evidence that the effectiveness gap of LLMs between English and Spanish is inversely correlated with model quality and size: smaller models have lower performance values in both languages and also a larger gap between their effectiveness in English and Spanish, and (iv) Empirical evidence that the relative performance of models on UNED-ACCESS 2024 is almost identical to their performance on an equivalent MMLU dataset and that results correlate with the public MMLU results, which may imply that contamination affects similarly to all language models, and that a small, sufficiently diverse dataset is representative enough to measure model performance. 

\section{Related work}
\label{s:related}

We summarize related work regarding current approaches to evaluate LLMs, existing benchmarks from human evaluations, multilingual LLMs evaluations and the problem of data contamination.

\paragraph{LLM benchmarking.} General benchmarking of the comparative performance of LLMs is currently performed either with datasets of questions and answers (most commonly in multiple-choice format) or with {\em LLM arenas} where users make requests, are offered answers from two or more LLMs, and pick their preferred one \citep{chatbotarena}. Here we focus on the first approach, where UNED-ACCESS 2024  belongs. 


Standard question/answer datasets include tasks such as common sense reasoning: HellaSwag \citep{zellers2019hellaswagmachinereallyfinish}, WinoGrande \citep{sakaguchi_winogrande_2019}, ARC~\citep{clark_think_2018},  OpenBookQA~\citep{mihaylov_can_2018}, CommonsenseQA~\citep{talmor-etal-2019-commonsenseqa}, PIQA~\citep{bisk2019piqareasoningphysicalcommonsense}, SIQA~\citep{sap-etal-2019-social}; world knowledge: TriviaQA \citep{joshi-etal-2017-triviaqa}, NaturalQuestions \citep{kwiatkowski-etal-2019-natural}; reading comprehension: RACE \citep{lai_race_2017}, QuAC \citep{choi2018quacquestionanswering} and SQuAC \citep{rajpurkar2016squad100000questionsmachine}; mathematical reasoning: MATH \citep{hendrycks2021measuringmathematicalproblemsolving}, GSM8K \citep{cobbe_training_2021}); and code generation: HumanEval \citep{chen2021evaluatinglargelanguagemodels}, MBPP \citep{austin2021programsynthesislargelanguage}. Aggregated benchmarks like MMLU \citep{hendrycks_measuring_2021}, AGIEval \citep{zhong_agieval_2023}, Big-Bench \citep{srivastava_beyond_2022}, and Big-Bench Hard \citep{suzgun2022challengingbigbenchtaskschainofthought} are also commonly used. In particular, since the introduction of GPT-3, evaluations that comprise a variety of datasets have been standardised (see for instance evaluations of GPT-3~\citep{brown2020languagemodelsfewshotlearners}, GPT-4~\citep{openai2024gpt4technicalreport}, Llama and Llama-2~\citep{touvron2023llama2openfoundation}, Mistral~\citep{jiang_mistral_2023} or Claude-3~\citep{claude_2024}). For Spanish and its official varieties, a new benchmark for the evaluation of open generative LLMs, named \textit{La Leaderboard}, has recently been released \cite{laleaderboard2024}.

Evaluations often mix prompting strategies, varying from zero-shot, few-shot with different numbers of examples, and chain-of-thought configurations. For example, in the evaluation of Llama-2 \citep{touvron2023llama2openfoundation} we find a zero-shot setting for evaluating common sense reasoning, except for one dataset in which 7-shot is used, and for aggregated benchmarks we find 3 and 5-shot settings. In more recent evaluations, such as that of Claude-3 \citep{claude_2024}, few-shot experimentation is performed with  different numbers of examples (varying from 0 to 25) and chain-of-thought strategies. 


\paragraph{Exams benchmarks.}  MMLU \citep{hendrycks_measuring_2021} is a popular benchmark with English multiple-choice questions covering subjects like mathematics, history, and computer science, ranging from elementary to college level. AGIEval \citep{zhong_agieval_2023} assesses human-like cognition and problem-solving abilities using high-stakes exams, such as university entrance and law school exams, in both English and Chinese. GPQA \citep{rein_gpqa_2023} presents challenging questions in biology, physics, and chemistry, aimed at testing even expert knowledge. M3Exam \citep{zhang_m3exam_2023} evaluates multilingual, multi-modal intelligence in LLMs using questions from human exams in nine languages. Additional benchmarks include GSM8K \citep{cobbe_training_2021} for elementary-level math, EXAMS \citep{hardalov-etal-2020-exams}, which covers 24 school subjects across 16 languages, OpenBookQA \citep{mihaylov_can_2018} and ARC \citep{clark_think_2018} for assessing scientific knowledge in English, and RACE \citep{lai_race_2017}, which focuses on reading comprehension from real English exams designed by Chinese teachers.

\paragraph{Multilingual evaluations.} Recent research has increasingly focused on multilingual evaluations of models like ChatGPT. For instance, \citet{lai-etal-2023-chatgpt} evaluated ChatGPT on 37 languages ranging from high to extremely low resources, across seven tasks in a zero-shot learning setting. Their findings, along with \citet{bang-etal-2023-multitask}, show that models struggle to generalize in low-resource languages. Similarly, M3Exam \citep{zhang_m3exam_2023} assesses models like GPT-4 across nine languages, revealing persistent difficulties in handling low-resource and non-Latin script languages. GPT-4's technical report itself extends its evaluation to a multilingual version of MMLU self-translated via Azure Translate \citep{openai2024gpt4technicalreport}, and a professional manual translation of MMLU in 14 languages (MMMLU) has recently been released \cite{mmmlu2024}.

To address multilingual challenges, benchmarks such as XTREME \citep{hu2020xtrememassivelymultilingualmultitask}, XTREME-R \citep{ruder2021xtremerchallengingnuancedmultilingual}, and MEGA \citep{ahuja_mega_2023} have been developed. Recently, a benchmark was used to assess general knowledge of the models, and compare English and Chinese \citep{gu2024xiezhieverupdatingbenchmarkholistic}. Additionally, models like Mixtral \citep{jiang2024mixtralexperts} include multilingual evaluations at launch, and works such as those of \citet{li2024quantifyingmultilingualperformancelarge} and \citet{blasi2021systematicinequalitieslanguagetechnology} highlight how model performance correlates with the amount of data available in each language during pretraining. 


\paragraph{Data contamination.} Machine learning models, especially LLMs, are increasingly dealing with the problem of data contamination, where models are trained on datasets from the internet, including those used for evaluation, leading to biased outcomes as the models may already know the answers beforehand \citep{Jiang2024InvestigatingDC, dong2024generalizationmemorizationdatacontamination, golchin2024timetravelllmstracing, sainz-etal-2023-nlp, yang2023rethinkingbenchmarkcontaminationlanguage}. This issue is gaining more attention, as seen in recent works and efforts like the LM Contamination Index\footnote{\url{https://hitz-zentroa.github.io/lm-contamination/}}. Several techniques have been proposed to detect contamination, such as checking dataset release dates or availability on the web, but these approaches may fail to address issues like model updates or indirect data leakage \citep{ahuja_mega_2023}. To mitigate this, researchers  suggest generating variations of datasets to assess reasoning capabilities \citep{srivastava_functional_2024}, or using private benchmarks to minimize contamination \citep{rajore2024truceprivatebenchmarkingprevent}, approach that we follow.

We introduce UNED-ACCESS 2024 to complement current evaluations by providing a bilingual and comparable resource featuring parallel questions in both Spanish and English. The questions are sourced from university private exams to minimize contamination, and have been professionally translated. Furthermore, by conducting experiments in a zero-shot mode, we aim to offer a more uniform evaluation compared to existing methods.

\begin{table*}[ht]
\centering
{\scriptsize 
\begin{tabular}{|p{1.7cm}|p{2cm}|p{1.2cm}|p{1.8cm}|p{2cm}|p{3cm}|p{1.2cm}|}
    \hline
     & \textbf{Languages} & \textbf{Original} & \textbf{Translations} & \textbf{Same instances per language} & \textbf{Translation method} & \textbf{Access}\\
     \hline
    \textbf{UNED-ACCESS 2024} & Spanish, English & Spanish & Yes & Yes & Professional manual translations, no external software & Private \\ \hline
    
    \textbf{AGIEval}  
    \citep{zhong_agieval_2023} & English, Chinese & English, Chinese & No translations, except in LogiQA & No, except in LogiQA & Manual transalation from Chinese & Public \\ \hline
    
    \textbf{Xiezhi} \citep{gu2024xiezhieverupdatingbenchmarkholistic} & Chinese, English & Chinese & Yes, from Chinese to English & Yes &  Google Translate API, followed by manual post-processing & Public \\ \hline
    
    \textbf{M3Exam} \citep{zhang_m3exam_2023} & English and 8 other languages & 9 languages & No & No & No translations & Public \\ \hline
    
    \textbf{EXAMS} \citep{hardalov-etal-2020-exams} & English, Spanish and 14 other languages & 16 languages & No, except for parallel questions & No, except for parallel questions in 7 languages & No translations & Public \\ \hline
    
    \textbf{XTREME} \citep{hu2020xtrememassivelymultilingualmultitask} and \textbf{XTREME-R} \citep{ruder2021xtremerchallengingnuancedmultilingual} & English, Spanish and 38 other languages, 9 tasks (XTREME); English, Spanish and 48 other languages, 10 tasks (XTREME-R) & Some tasks$^{*}$ & Some tasks & Some tasks & Some professional translations, some parallel instances and some automatic translations$^{**}$ & Public \\ 
    \hline
    
    \textbf{MEGA} \citep{ahuja_mega_2023} & English, Spanish and 68 other languages (16 datasets) & Some tasks & Some tasks & Some tasks & Some datasets contain professional translations, parallel instances and automatic translations$^{***}$ & Public \\
    \hline

    \textbf{MMMLU} \citep{mmmlu2024} & English, Spanish and other 13 languages & English & Yes & Yes & Professional human translations & Public \\
    \hline
\end{tabular}
}
\caption{Comparison of translation methods in multilingual datasets. ~$^{*}${While in the XNLI \citep{conneau-etal-2018-xnli} or XQuAD \citep{artetxe-etal-2020-cross} datasets the translations from English were done manually by professional translators, in others such as MLQA instances exist naturally in several languages, without relying on English translations.} ~$^{**}${Since not all tasks are available in all languages, to enable a broader comparison across languages they automatically translate some English instances to the remaining languages using an in-house translation system.} ~$^{***}${In order to compare diverse prompting strategies, they use Bing translator to translate target language data to English.}}
\label{traducciones}
\end{table*}

\section{The UNED-ACCESS 2024 Dataset}
\label{dataset}

UNED-ACCESS 2024 contains 1003 multiple-choice questions, with 3 or 4 possible answers, from the following subjects of the UNED Access Course for Over 25s: Business Administration and Management, Biology, Biochemistry, Economics, Fundamentals of Computer Science, Spanish Language, Literature, Mathematics, Mathematics Applied to Social Sciences, Advanced Mathematics and Psychology. The questions, originally in Spanish and human-translated into English, are from exams conducted between 2009 and 2022. Although several benchmarks of multiple-choice questions exist, some of them, such as ARC~\citep{clark_think_2018} and OpenBookQA~\citep{mihaylov_can_2018} have been found to be limited as they cover simple topics of primary school subjects, for which models can already achieve solid performance. UNED-ACCESS 2024 covers more complex topics and questions that require from simple memorization to mathematical reasoning or linguistic skills.  Table \ref{traducciones} shows a comparison of several multilingual datasets and benchmarks, regarding whether they are public, the languages included, the original language of the questions or instances, and the availability of translations and the translation method used. It shows that UNED-ACCESS 2024 is the only dataset with private access and fully manual translation.

Appendix \ref{sec: apendice dataset} contains more details on the compilation process, the format of the dataset, and the dataset statement. The following are two examples of questions in English:

{\footnotesize
\begin{itemize}
\setlength\itemsep{0,1em} 
    \item {\bf Spanish Language}: \newline Which of the following sentences uses accents correctly?  A. Es por éso.  B. Fui y volví.  C. Llorar y reir.  D. No se prohibe.


    \item {\bf Mathematics Applied to Social Sciences}: \newline When \textbackslash(x \textbackslash to 1\textbackslash), the function \textbackslash(f(x) = \textbackslash frac\{1-x\^{}2\}\{1-x\}\textbackslash): A. Has a limit of \textbackslash(\textbackslash infty\textbackslash).  B. Has a limit of 2.  C. Has no limit.
\end{itemize}
}

Table \ref{tabla_distribucion} shows, per subject, the  number of questions, the total word count and the number of answer options per question, which is relevant since the probability of getting the answer right randomly depends on the number of possible answers. 

\begin{table}[ht]
\centering
\small
\begin{tabular}{p{.22\textwidth}ccc}
\hline
\textbf{Subject} & \textbf{\# Q}  & \textbf{\# AxQ}  & \textbf{\# W} \\
\hline
Business and Administration Management & 87  & 3 & 3849 \\
Biology & 119  & 3 & 2753 \\
Biochemistry & 59 & 3 & 1407 \\
Economics & 51  & 4 & 1725 \\
Foundations of Computer Science & 63  & 4 & 1987 \\
Spanish Language & 94  & 4 & 2816 \\
Literature & 91  & 4 & 5130 \\
Mathematics & 73  & 3 & 1465 \\
Mathematics Applied to Social Sciences & 94  & 3 & 2847 \\
Advanced Mathematics & 24  & 3 & 446 \\
Psychology & 248  & 4 & 5669 \\
\hline
Total & 1003 & -- & 30094 \\
\hline
\end{tabular}
\caption{Distribution of the number of questions per subject  (\# Q), the number of answer options per question (\# AxQ) and total word count (\# W) per subject in the Spanish UNED-ACCESS 2024 dataset.}
\label{tabla_distribucion}
\end{table}

\section{Experimental setup}
\label{experim_setup}


Experiments were conducted using 12 generative models: four proprietary models (GPT-4-Turbo \citep{openai2024gpt4technicalreport}, GPT-4o, GPT-3.5-Turbo \citep{brown2020languagemodelsfewshotlearners} and Claude-3-Opus \citep{claude_2024}) and eight open-weights models: Llama-2-7B \citep{touvron_llama_2023}, Llama-3-8B \citep{llama_3_2024}, Llama-3-70B, Gemma-7B \citep{gemmateam2024gemmaopenmodelsbased}, Gemma-2-27B, Mixtral-8x7B \citep{jiang2024mixtralexperts}, Mistral-7B \citep{jiang_mistral_2023} and Leniachat-Gemma-2B, all trained for instruction following. In Appendix \ref{appendix-model access} details are provided about how the models were accessed.


We decided to apply a uniform zero-shot setting because it more closely resembles the real-world scenario in which people interact with LLMs, and it also results in a simpler and more replicable experimentation. As for the hyperparameters, the temperature was set to 0 in all models to ensure deterministic responses, minimizing creativity and focusing on factual or reasoning-based answers, which suits the dataset. The questions were provided one at a time and the prompt is fixed, providing a system prompt, a user prompt and an assistant prompt. As extra information, the prompt provided the name of the subject to which the question belongs. The language of the prompt was always in the same language of the dataset used to evaluate the model (English or Spanish) as in  \citet{zhang_m3exam_2023} and \citet{openai2024gpt4technicalreport}. The prompts are the following for Spanish (ES) and English (EN):



{\small
\begin{itemize}
\setlength\itemsep{0,01em} 
    \item {\bf System prompt}\newline {\bf ES:} \texttt{Eres un sistema experto en responder preguntas de exámenes.} \newline {\bf EN:} \texttt{You are an expert system for answering exam questions.}
    \item {\bf User prompt}
    \newline {\bf ES:} \texttt{Responde a la siguiente pregunta de la asignatura \{\}, tan solo con la letra de la respuesta correcta. Pregunta: {} }
    \newline {\bf EN:} \texttt{Answer the following question of the subject \{\} only with the letter of the correct answer. Question: {} }
    \item {\bf Assistant prompt}
        \newline {\bf ES:} \texttt{Letra de la respuesta correcta:}
    \newline {\bf EN:} \texttt{Letter of the correct answer: }
\end{itemize}
}

Additionally, for open models the instruction was formatted according to the labels used in training, as explained in their model cards. These templates are shown in Appendix \ref{sec:appendice instruct}. Finally the literal output of the models was cleaned up and structured before applying the evaluation script, since, on many occasions, the answer contained answer justifications or additional material. The letter of the correct answer was kept if provided.

\paragraph{Evaluation metrics.} In the literature, evaluation is typically based on Accuracy, the proportion of correct answers ($C$) over the total responses ($N$). However, to allow for direct comparisons across subjects with varying numbers of answer choices, we use Cohen’s Kappa coefficient, which accounts for the number of possible choices ($M$) in multiple-choice questions:

\vspace{-0.5cm}
\begin{equation*}
\scriptsize
    \text{Kappa} = \frac{\text{observed accuracy} - \text{expected accuracy}}{1 - \text{expected accuracy}} = \frac{\frac{\text{C}}{\text{N}} - \frac{1}{\text{M}}}{1 - \frac{1}{\text{M}}}
\end{equation*}

where the expected accuracy corresponds to that of random guessing: 1/3 for questions with 3 options and 1/4 for questions with 4 options. Cohen's Kappa adjusts the correctness level so that random answers result in a Kappa of zero, allowing comparison across exams with varying answer choices. Kappa values range from -1/2 to 1, with negative values indicating worse-than-random performance. The final result for each model and language is the arithmetic mean of Kappa values across all subjects, giving equal weight to each subject to account for dataset imbalances.

\section{Results}
\label{results}

\begin{table}[ht]
    \centering
    \resizebox{\columnwidth}{!}{
    \begin{tabular}{lcccr}
    \toprule
    \textbf{Model} &  \textbf{Type} & \textbf{EN} & \textbf{ES} & \textbf{Gap\%} \\
    \midrule
    Claude-3-Opus &  proprietary & 0.79 & 0.81 & -2.51 \\
    GPT-4o &  proprietary & 0.78 & 0.77 & 0.83 \\
    GPT-4-Turbo &  proprietary & 0.76 & 0.78 & -1.72 \\
    Llama-3-70B-Instruct & open & 0.65 & 0.67 & -3.44 \\
    Gemma-2-27B-Instruct & open & 0.64 & 0.66 & -3.40 \\
    GPT-3.5-Turbo & proprietary & 0.60 & 0.55 & 9.19 \\
    Mixtral-8x7B-Instruct & open & 0.56 & 0.57 & -2.32 \\
    Llama-3-8B-Instruct & open & 0.51 & 0.50 & 2.73 \\
    Mistral-7B-Instruct & open & 0.46 & 0.43 & 5.67 \\
    Gemma-7B-It & open & 0.41 & 0.38 & 8.93 \\
    Llama-2-7B-Chat & open & 0.32 & 0.25 & 26.04 \\
    Leniachat-Gemma-2B & open & 0.15 & 0.11 & 37.27 \\
    \bottomrule
    \end{tabular}
    }
    \caption{Average performance of each model on UNED-ACCESS 2024 (Cohen's Kappa) and language gap \%. Results are sorted by performance in English.} 
    \label{tabla_gap}
\end{table}


Table \ref{tabla_gap} shows the average Cohen's Kappa results by model in Spanish and English. Additional detailed results per language, model and subject are provided in Table \ref{base_asig} in Appendix \ref{sec:appendix-results}. The best overall results for Spanish are achieved by the proprietary models Claude-3-Opus, GPT-4-Turbo and GPT-4o, followed by the open models Llama-3-70B, Gemma-2-27B and Mixtral-8x7B, which outperform GPT-3.5-Turbo.  In the pass margin (around 0.50 points) there is Llama-3-8B, and below are Mistral-7B, Gemma-7B, Llama-2-7B and Leniachat-2B. For English, the  ranking of the models by average performance is almost identical to that of  Spanish, except for an exchange in the positions of GPT-3.5 and Mixtral (GPT-3.5 performs better than Mixtral in English but worse in Spanish). At the quantitative level, the results are also very similar in English and Spanish. 

\paragraph{EN-ES Language gap.} 
The last column in Table \ref{tabla_gap} shows the performance gap between English and Spanish. We calculate the percentage effectiveness gap between languages from the averaged results as:  

{\footnotesize
\begin{equation*}
    \text{GAP\% EN-ES} = \frac{\text{(EN-ES)}}{\text{ES}}\cdot 100
\end{equation*}
}



Five out of twelve models perform slightly better in Spanish than English (namely Claude-3, GPT-4, Llama-3-70B, Gemma-2-27B and Mixtral-8x7B, where the gap is negative), two of them proprietary (which perform the best in Spanish overall) and three open (also the three best models among the open-weights ones). This does not imply that these models are better in Spanish: English questions are manual translations of the original Spanish ones, and  there are at least three language-related factors that affect overall performance: (i) even with high-quality manual translations, there might be translation artifacts (e.g. unusual translation choices) that make the English text harder to process; (ii) contamination: even if the questions are private, the chance that some questions have been seen by the models in Spanish is higher than in English; (iii) language competence of the models in English vs Spanish. Presumably, in the models that perform better in Spanish, effects (i) and (ii) out-weight pure language competence. 


Among the proprietary models, three of them have similar performance differences (in absolute value), whilst that of GPT-3.5 is the most pronounced, in favour of English. As for the open models, the two best performing ones (Llama-3-70B and Gemma-2) have a very similar performance difference, both in favour of Spanish (and larger that the ones observed for the best proprietary models). Mixtral is the next model in the ranking and still favours Spanish, while all the other models, which do not pass (or are at the limit), favour English. In this set of models the performance difference is very pronounced, specially for Llama-2 and Leniachat, which are the worst performing models. In summary, for the best models we observe a slight advantage for Spanish, while in the worst models there is a pronounced advantage towards English. The best models are in turn larger, and therefore more likely to have been trained with more Spanish data, and more data from the web in general, whereas smaller models may rely more solely on English. However, in principle this is not the case of {\em Leniachat}, which has been conceived specifically for Spanish. The size of the models is crucial: although we do not know for certain the size of the proprietary models, in general model size is the best performance predictor, with one exception: the very similar performance between Llama-3-70B and Gemma-2-27B, with such a large difference in size, is noteworthy.

There are two cases where the gap is extremely pronounced (in favour of English), coinciding with the two worst-performing models: Leniachat, which is the smallest model, and Llama-2, which is the oldest of the evaluated models.

There is a strong, \textbf{statistically significant inverse correlation between model quality and language gap}: Pearson's correlation between English performance and EN-ES gap is -0.87 with a p-value of 0.0002; and Pearson's correlation between performance in Spanish and language gap is -0.89 (p=9.83e-05). The performance of less powerful models degrades faster in Spanish than in English. 

\begin{figure*}[htb] 
    \centering 
    \includegraphics[width=\textwidth]{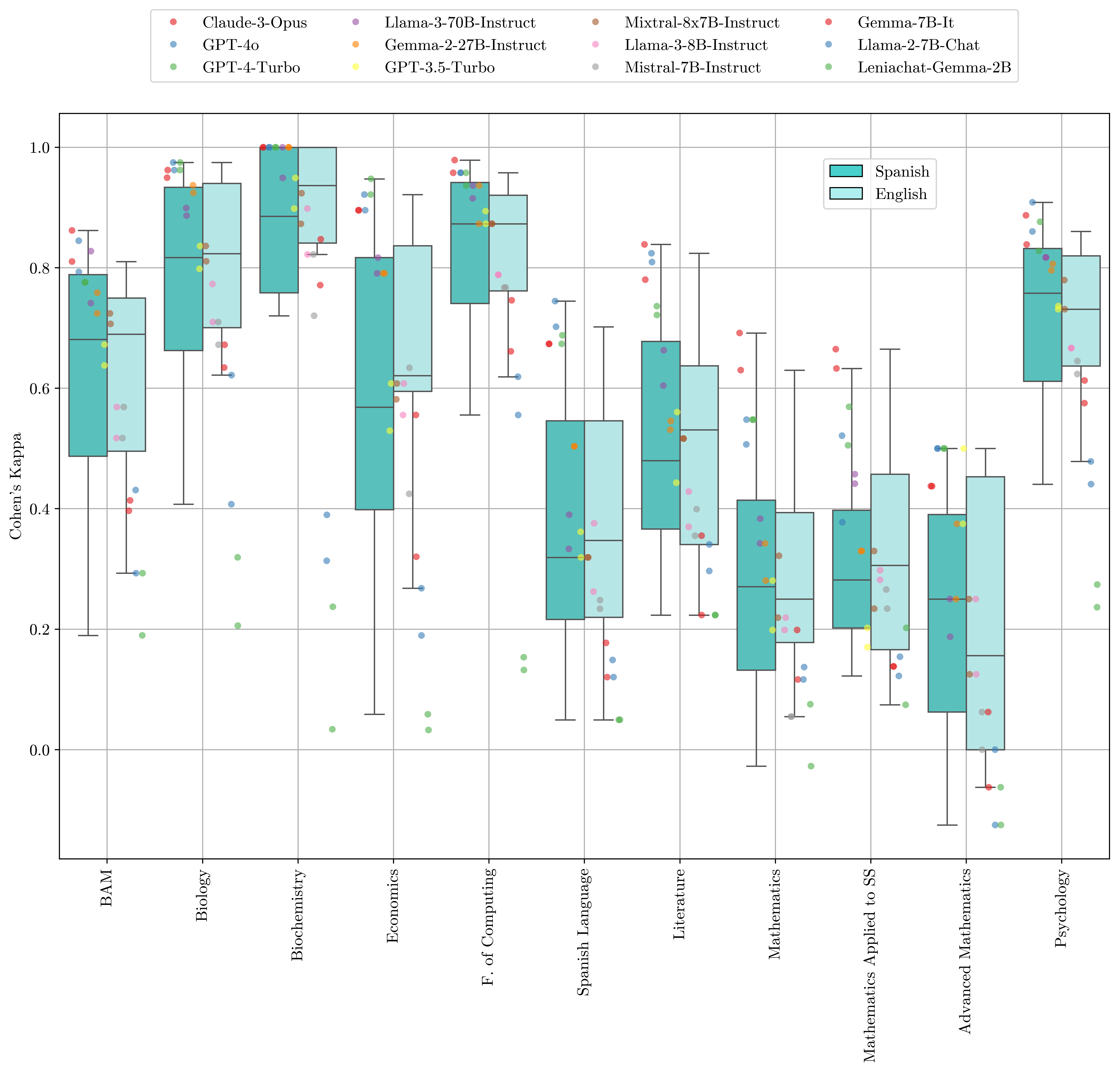}
    \caption{Distribution of results on UNED-ACCESS 2024  per subject (Cohen's Kappa).} 
    \label{model-subject} 
\end{figure*}

\paragraph{Results by subject.} Figure \ref{model-subject} shows the performance per subject for all models in both languages (also provided in Table \ref{base_asig} in Appendix \ref{sec:appendix-results}). Subjects with the highest scores tend to be Biology, Biochemistry, Fundamentals of Computer Science and Psychology. In Economics and Business Administration and Management (BAM), a slight decrease in scores is observed, as well as in Literature and Spanish Language. The worst results are observed in the three Mathematics subjects, with slightly higher results in Mathematics Applied to Social Sciences, which is presumably the least complex of the three. Just as the low performance in Mathematics subjects can be explained by the reasoning abilities that the questions require, the lower results in Spanish Language may also be due to the fact that the models are trained on tokens but do not really distinguish a letter or a word, and therefore fail in tasks that require concrete lexical, syntactic or orthographic understanding. Besides, the fact that the models solve Biochemistry questions so well in both languages suggests that these questions only involve recalling fundamental knowledge; that the information  needed to answer the questions is well represented in internet;  that the questions do not require complex reasoning or deep contextual interpretation; and on the other hand, that the models understand the formulas expressed in \LaTeX. This is a hint that formulas in \LaTeX\ should not be what prevents models from solving the Math questions correctly.


Broadly speaking, the best models perform best in all subjects and, conversely, the worst models remain at low scores throughout the dataset. The best performing model in average, Claude-3, passes all subjects except for Advanced Mathematics, suggesting that even the most advanced models do not show true reasoning capabilities as an emergent phenomenon. Claude-3 scores above 0.8 in seven out of the eleven subjects in Spanish (BAM, Biology, Biochemistry, Economics, Fundamentals of Computer Science, Literature and Psychology) and six in English. In Spanish Language it remains at 0.67 in both languages, and around 0.65 in Mathematics and Mathematics Applied to Social Sciences. The results for GPT-4 and GPT-4o are very similar. Llama-3-70B, Gemma-2-27B and GPT-3.5 follow the lead, with results that drop overall by at least one tenth with a significant drop in the three Math subjects. Results continue to fall with Mixtral-8x7B, GPT-3.5 and Llama-3-8B, which fall outside the pass mark. The remaining models have results that hover around random probabilities, in some cases performing even worse than a random answer (values below zero), with the lowest results generally obtained with Gemma-7B, Llama-2-7B and Leniachat-Gemma-2B. It is interesting to note that although Claude-3 achieves the highest average result, in numerous subjects it is GPT-4 or GPT4o models that achieve the highest results.

\paragraph{UNED-ACCESS vs MMLU: public vs private test sets.}  A direct comparison between the public results on MMLU \citep{hendrycks_measuring_2021} and UNED-ACCESS 2024 is not adequate: the dataset sizes are different (MMLU has more than 14000 test instances) and the subjects and levels differ: MMLU ranges in difficulty from an elementary level to an advanced professional level, whereas UNED-ACCESS 2024 covers only university-entrance level. Also, most public evaluations of MMLU are 5-shot and even include chain of thought prompting, as compared to the zero-shot setting of our experiments.  For these reasons we have performed our own evaluation of the models, using the same methodology described earlier, and selecting the subset of 2822 MMLU questions (\textit{MMLU-10}) from ten topics which correspond most closely to the level and topics of UNED-ACCESS: high school biology, high school chemistry, high school computer science, elementary mathematics, high school mathematics, high school physics, high school macroeconomics, high school microeconomics, high school psychology and high school world history. 

\renewcommand{\arraystretch}{1.5}
\begin{table}[ht]
    \centering
    \resizebox{\columnwidth}{!}{
    \begin{tabular}{@{}lcc@{}}
    \toprule
     & \makecell{\textbf{MMLU} \vspace{-1mm}\\ \small{from publications}} & \makecell{ \textbf{MMLU-10} \vspace{-1mm}\\ \small{from zero-shot experiments}}  \\
    \midrule
    GPT-4o & 88.7 & 83.40 \\
    Claude-3-Opus  & \makecell{86.8\vspace{-1mm}\\ \tiny{(5-shot)}}  & 83.71 \\
    GPT-4-Turbo  & \makecell{86.5} & 81.61 \\
    Llama-3-70B-Instruct  & \makecell{82.0 \vspace{-1mm}\\ \tiny{Llama-3-70B (5-shot)} } & 75.47 \\
    Gemma-2-27B-Instruct  & \makecell{75.2 \vspace{-1mm}\\ \tiny{Gemma-2 (5-shot)}} & 76.61 \\
    Mixtral-8x7B-Instruct  & \makecell{70.6 \vspace{-1mm}\\ \tiny{Mixtral 8x7B (5-shot)} }  & 65.34 \\
    GPT-3.5-Turbo  & 70.0 & 63.74 \\
    Llama-3-8B-Instruct  & \makecell{ 68.4 \vspace{-1mm}\\ \tiny{Llama-3-8B (5-shot)}} & 61.54 \\
    Mistral-7B-Instruct & \makecell{58.4 \vspace{-1mm}\\ \tiny{(5-shot)}} & 54.24 \\
     Gemma-7B-It  & \makecell{53.3 \vspace{-1mm}\\  \tiny{(5-shot)}} & 42.50 \\
    Llama-2-7B-Chat  & \makecell{45.3 \vspace{-1mm}\\ \tiny{Llama-2-7B (5-shot)}}  & 35.08 \\
    Leniachat-Gemma-2B  & - & 29.81 \\
    \bottomrule
    \end{tabular}
    }
    \caption{{\bf Accuracy} scores of models obtained from published MMLU results in their respective reports, and  results obtained from experiments on MMLU-10 executed with the experimental setting described in this paper.}
    \label{mmlu}
\end{table}

First, we have checked whether the results that we obtain on MMLU-10 with our methodology correlate with the published results.
Table \ref{mmlu} provides the MMLU results published in the official benchmark\footnote{\url{https://paperswithcode.com/sota/multi-task-language-understanding-on-mmlu}} (or, in their absence, in the respective model reports) and the results of applying our methodology on MMLU-10. Both evaluations are reported
 in terms of Accuracy, rather than Cohen's Kappa, because published results are always presented with this metric. This is the only table in this paper that presents results using Accuracy.
Pearson's correlation between these two sets of results is 0.9863 with a p-value of 2.31e-08, which means that results are highly correlated. This high correlation validates our experimentation, as it implies that our choices of prompts and hyperparameters, and zero-shot setting, yield results consistent with those found in the literature.

This analysis allows us to compare the  results obtained on MMLU-10 with the results on UNED-ACCESS 2024 in English and check whether they correlate.  Table \ref{acceso_mmlu} shows both results in terms of Cohen's Kappa (recall that the MMLU questions all have 4 answers). Pearson's correlation between these results is 0.98 (p=5.04e-08): both results have almost perfect, statistically significant correlation. This consistency reinforces the reliability of results obtained with both datasets.


\renewcommand{\arraystretch}{1}
\begin{table}[!ht]
    \centering
    \resizebox{\columnwidth}{!}{
    \begin{tabular}{@{}lcc@{}}
    \toprule
     & \makecell{\textbf{UNED-ACCESS 2024}  \vspace{-1mm}\\ \tiny{EN, 1K questions}} & \makecell{ \textbf{MMLU-10} \vspace{-1mm}\\ \tiny{EN, 2.8K questions}} \\
    \midrule
    Claude-3-Opus & 0.79  & 0.78 \\
    GPT-4o & 0.78 & 0.78 \\
    GPT-4-Turbo & 0.76  & 0.75 \\
    Llama-3-70B-Instruct & 0.65 & 0.67 \\
    Gemma-2-27B-Instruct & 0.64 & 0.69 \\
    GPT-3.5-Turbo & 0.60 & 0.52 \\
    Mixtral-8x7B-Instruct & 0.56 & 0.54 \\
    Llama-3-8B-Instruct & 0.51 & 0.49 \\
    Mistral-7B-Instruct & 0.46 & 0.39 \\
    Gemma-7B-It & 0.41 & 0.23 \\
    Llama-2-7B-Chat & 0.32 & 0.13 \\
    Leniachat-Gemma-2B & 0.15 & 0.06 \\
    \bottomrule
    \end{tabular}
    }
    \caption{Cohen's Kappa results of models on UNED-ACCESS 2024 (English), and results from experiments on a subset of the MMLU executed with our prompts and hyperparameters in a zero-shot setup.}
    \label{acceso_mmlu}
\end{table}

The fact that the results on both datasets are so similar, in spite of the fact that UNED-ACCESS should have minimal contamination, implies that either the questions in UNED-ACCESS are somewhat easier in average, or that there is some kind of indirect contamination: some questions might be recurring for some subjects, and models might have seen variations of the questions even if they do not have access to the original ones. Also, the similarity of the results seems to indicate that the translation effect is not decisive. 

In Table~\ref{t:MMLU-subjects} of Appendix~\ref{apendice_mmlu10} we also provide the evaluation results on the  MMLU-10 dataset per subject, which shows the same trends as for UNED-ACCESS 2024. The worst results are obtained for Mathematics and the best results are obtained for Biology, Micro- and Macro-economics, Computer Science, World History and Psychology. Also equally, the best models perform best in all subjects and the best performing models in average, GPT-4o and Claude-3, pass all subjects except for Mathematics.

\section{Conclusions}
\label{s:conclusions}


We introduced the UNED-ACCESS 2024 bilingual dataset to evaluate general knowledge of language models. It consists of 1003 private multiple-choice questions and answers in Spanish, manually translated into English. We evaluated proprietary and open-source LLMs on this dataset and on an equivalent subset of MMLU (MMLU-10), with the same methodology, addressing three research questions.

\textbf{RQ1: What is the performance of models in a dataset with minimal contamination?} We have found that the behaviour of models is almost identical on our dataset and on MMLU-10: the model rankings provided by both datasets are almost equivalent (Pearson correlation is 0,98). In both datasets, performance is strongly correlated with model size. Across subjects, all models perform substantially worse in subjects requiring reasoning and calculus, compared to those which require mainly memorization. We plan to classify the questions according to the level of memorisation or reasoning they require, as in \citet{yu2023kolacarefullybenchmarkingworld} where they use Bloom's taxonomy of educational objectives~\citep{bloom1956taxonomy} in order to better assess models on these dimensions. Overall, the fact that UNED-ACCESS has minimal contamination does not lead to any meaningful difference in the relative models' performance. This suggests that contamination affects similarly to all models, or that there are indirect sources of contamination in our dataset. In a preliminary investigation, however, a web search for solved exams in each of the subjects did not lead to any systematic repository with questions and answers, although we occasionally found PDF versions of the exams with the right answers highlighted with a pen (usually from academies that prepare the students for the exams). 

\textbf{RQ2: What is the performance gap of models between English and Spanish?} Our most significant finding is that the effectiveness gap between English and Spanish is inversely correlated with model accuracy: the best models have smaller relative differences between their performance in both languages. In general, smaller models show larger gaps, and the best models even perform slightly better in Spanish than English. The original questions are in Spanish, and therefore this negative gap may indicate that the true language gap (most likely in favour of English) is out-weighted by translation effects and contamination (even if small, it is more likely in Spanish as it is the source language). More experimentation is needed to isolate these factors from each other. 

\textbf{RQ3: How suitable is the zero-shot approach to evaluate LLMs on 1,000 exam questions?} 
The high correlation observed between the results in MMLU and UNED-ACCESS 2024 in English, along with the high correlation between Spanish and English results in UNED-ACCESS 2024, suggest that (i) our experimental setting is fit for reliable experimentation, since it yields results consistent with those found in the literature, and (ii) despite its reduced size, our dataset is sufficiently diverse and representative to measure performance by model and discipline, and the effect of manual translation (Spanish to English) does not significantly alter results.  

To further mitigate contamination and increase difficulty, we plan to extend the dataset with undergraduate-level questions of increasing difficulty, and also to check the real level of contamination to which they may be exposed. We also plan to estimate the consistency of models' answers to prompt variations, that is, how the models deal with changes in content (and get to measure a \textit{reasoning gap} \citep{srivastava_functional_2024}), and in form (to measure the \textit{performance spread} \citep{polo2024efficientmultipromptevaluationllms, sclar2023quantifyinglanguagemodelssensitivity, mizrahi2024stateartmultipromptllm}), and apply this methodology to other benchmarks. 


Finally, although the questions and answers will not be distributed to keep contamination minimal, new relevant open models will continue to be evaluated in a UNED-ACCESS leaderboard at \url{https://leaderboard.odesia.uned.es}. Developers with an open model that wish to be included in the evaluation may send a request to \texttt{odesia-comunicacion@lsi.uned.es}.

\section*{Limitations}

Firstly, although the correct answers of the exam questions of UNED-ACCESS 2024 have never been made public officially, which makes us believe that the risk of {\bf contamination} is limited, we have not measured the real level of contamination by, for example, making systematic searches in the internet in order to determine whether and in what form some correct answers may be publicly available. As stated in the paper, more research is needed to reliably estimate contamination. 
Secondly, the {\bf size} of the dataset might be considered small as compared to the most popular public datasets. We have given priority to gather quality questions with private correct answers, rather than to build a large dataset. However, we are working towards producing a larger dataset with university level questions in a diversity of subjects. 
Third, our estimation of the \textbf{language gap} is just a first approximation, as there are confounding factors (translation, contamination) that are difficult to isolate from linguistic performance. More research is needed to find adequate methodologies to estimate the language gap. Also, our experimentation is directed at estimating a minimal gap (comparing English with other predominant online language such as Spanish); obviously, the gap will be larger with less resourced languages. 
Finally, we have not {\bf categorized} the questions into levels of cognitive complexity, for example using Bloom's taxonomy \citep{anderson2001taxonomy}, in order to make an in-depth study of the models performance per cognitive level. This will be addressed in future work. 

\section*{Acknowledgments}

This work has been funded by the European Union - NextGenerationEU through the `Recovery, Transformation and Resilience Plan', by the 
{\em Ministerio para la transformación digital y de la función pública} and by UNED via cooperation agreement C039-21OT. 
However, the views and opinions expressed are solely those of the author(s) and do not necessarily reflect those of the European Union or the European Commission. Neither the European Union nor the European Commission can be held responsible for them.

\bibliography{references}

\appendix

\newpage
\section{UNED-ACCESS 2024: Additional Information}
\label{sec: apendice dataset}

This appendix provides more details about the collection, transcription and annotation process of the dataset UNED-ACCESS 2024, as well as an example of the format of the dataset instances, and the dataset statement.

\subsection{Data Harvesting and Preparation}
\label{sec:appendix details}

\paragraph{Collection of the exams.} 
 The exams have been downloaded directly from the UNED repository of exams, accessible for UNED users at \url{blind link}. The repository contains exams for Bachelor's, Master's, Languages, and Access Courses for 25 and 45 year olds. In this first version of the dataset, only exams from the Access courses for over 25s have been used, as they are a smaller set with a wide distribution of subjects. The availability of exams varies from subject to subject and, in particular, the availability of multiple-choice exams depends on the subject in question (some subjects have only, or mostly, essay questions). In the case of some subjects, such as Advanced Mathematics, no more questions were available than those included in the dataset. The exams have been downloaded between October and November 2023 and the questions obtained correspond to exams between 2009 and 2022.

\paragraph{Sampling and filtering.} Questions selected meet the following two conditions:

\begin{itemize}[itemsep=0pt, parsep=0pt]
    \item Availability of private correct answers. We have selected only exam questions for which templates with ``official'' solutions  were provided to us by UNED. These templates exist for a subset of the exams contained in the repository. Thus, we first filter out examsfor which no templates are provided.
    
    \item Multiple-choice questions. Once the exams for which a solution template is available have been filtered, multiple-choice exam question have been selected. Not all subjects are evaluated with this type of exams.
\end{itemize}

\paragraph{Preprocessing.} The first step to create the dataset consisted in matching the PDF exams with their corresponding answer templates provided by UNED. Exams available in the private repository of the UNED were in PDF format, so they have been transcribed using Nougat OCR\footnote{\url{https://facebookresearch.github.io/nougat/}} in Python. This OCR system was chosen because it is open-source and based on Transformers \citep{vaswani2023attentionneed}, and allows the transcription of formulas in \LaTeX, a functionality especially useful for the transcription of exams in subjects such as Mathematics or Biochemistry.  After conversion, the exams were structured into  CSV format with a Python script and questions and answers were matched. Due to variations in exam formats, each exam required individual treatment. Finally, duplicate questions, chained questions, questions with images, and those marked with an `X' in the answer template were removed. The original Spanish questions were then manually translated into English by a professional translator,  who was instructed not to use any machine translation tool nor generative models such as ChatGPT. Doubts about the translation of terminology were discussed with the authors of the paper. The translator was also instructed to work locally and to delete the files once the work was finished.

Subsequently, the plain text questions have been structured in CSV and JSON format semi-automatically by means of a Python script using libraries such as \texttt{re} to identify regular expressions. Thus, we have a bilingual dataset with the following fields:

\begin{itemize}[itemsep=0pt, parsep=0pt]
    \item \texttt{\textbf{question}}. Contains the statement of the question.
    
    \item \texttt{\textbf{A}}. Contains the text of answer option A.
    
    \item \texttt{\textbf{B}}. Contains the text of answer option B.
    
    \item \texttt{\textbf{C}}. Contains the text of answer option C.
    
    \item \texttt{\textbf{D}}. Contains the text of the answer option D, and in case it does not exist it is empty (NaN in CSV format or null in JSON format).
    
    \item \texttt{\textbf{solution}}. It contains the letter of the correct answer, extracted from the solution templates.
    
    \item \texttt{\textbf{year}}. Year of the examination.
    
    \item \texttt{\textbf{test\_name}}. This is the name of the PDF file of the exam, which contains information such as the subject code, year, month and week of the exam.
    
    \item \texttt{\textbf{code}}. Subject code.
    
    \item \texttt{\textbf{subject}}. Name of the subject.
    
    \item \texttt{\textbf{id}}. A unique identifier assigned to each question, consisting of the name of the test (test\_name) and a number indicating the order of the question within the subject. This identifier matches that of its corresponding English translation.
    
\end{itemize}

\subsection{Dataset Statement}

The characteristics of the dataset are listed in the form of a  dataset statement \citep{bender-friedman-2018-data}:

\begin{itemize}
\setlength\itemsep{0,1em} 
\item {\bf Type of instances}: Exam questions.
\item {\bf Type of questions}: Most questions consist  on choosing the correct answer, and some consist on filling the blank/s in the statement (see example in Appendix \ref{ejemplo_json}). Literature and Psychology questions primarily test encyclopedic knowledge. Spanish Language questions range from testing spelling, lexical, syntactic and grammatical knowledge. Business and Administration Management and Economics questions cover basic financial, accounting, and business management concepts, and few calculus questions. Fundamentals of Computer Science's are mostly memorization questions, although some require to perform some calculations. Biology questions focus mostly on concepts and definitions, with some of them requiring some reasoning, for example when applying genetics concepts. Biochemistry questions require mainly memorization, and many of them include chemical formulas. Mathematics Applied to Social Sciences questions contain basic analysis, statistics, logic, probability and other arithmetic problems. Mathematics questions cover analysis, algebra, combinatorics, probability, and others, and so does Advanced Mathematics with questions in a more advanced level of calculus, algebra, and trigonometry. Most of the mathematics questions include equations, polynomials or sets expressed in \LaTeX\ formulas, and require reasoning to solve the problems. 

\item {\bf Institution providing exams}: UNED, Spain.
\item {\bf Task}: Multiple-choice Question Answering.
\item {\bf Curation rationale}: Obtaining multiple choice questions to automatically evaluate LLMs, without diagrams or images.
\item {\bf Domain}: Academic.
\item {\bf Difficulty level}: University entrance.
\item {\bf Knowledge areas}: Business Administration and Management, Biology, Biochemistry, Economics, Fundamentals of Computer Science, Spanish Language, Literature, Mathematics, Mathematics Applied to Social Sciences, Advanced Mathematics and Psychology.
\item {\bf Languages and varieties}: Standard Spanish and Standard English.
\item {\bf Number of questions}:  1003 per language.
\item {\bf Year of publication of the exams}: From 2009 to 2022.
\item {\bf Total word count}:{\bf Spanish}: 30094; {\bf English}: 28186.
\item {\bf Partitions}: Test.
\item {\bf Formats}: CSV, JSON.
\item {\bf Year of creation of the dataset}: 2024.
\item {\bf Further considerations}:

\begin{itemize}
    \item The selected exams are a partial sample of the total number of exams available in the repository. 

    \item The dataset is self-contained and needs no external sources.

    \item The dataset is of small size (1003 questions) designed as a test set to evaluate generative models in a zero-shot setting, without model training.
    
    \item The dataset has been manually checked to minimize spelling, OCR, and \LaTeX\ errors. Duplicates, image-based questions, and those dependent on other answers have been removed.

    \item The dataset is kept private as it includes UNED exams with solution keys, to prevent answers from becoming public and contaminating generative models.

    \item Offensive/sensitive data: none.

    \item The dataset will remain private and won't be distributed, as it's intended for use in the ODESIA Benchmark for evaluating language models in English and Spanish.

    \item No maintenance of the dataset is foreseen.

    \item Extensions of the dataset are foreseen, possibly with undergraduate exams in several disciplines. 
    
\end{itemize}
\end{itemize}

\subsection{Format of the Dataset}
\label{sec:ap-dataset}


The dataset is presented in both CSV and JSON format, structured into the fields:  \texttt{question, (option) A, (option) B, (option) C, (option) D, solution, year (of exam), test name, code, subject, id}. Below is an example of a dataset object in JSON format where we omit the letter of the correct solution:

\begin{center}
\begin{minipage}{7cm}
\begin{lstlisting}[language=json]
{
    "question": "The empirical testing procedure, which is carried out by means of ... is called the ... method.",
    "A": "deducing conclusions; experimental",
    "B": "experimentation; experimental",
    "C": "the correlation of variables; experimental",
    "D": "experimentation; correlational",
    "solution": "correct answer",
    "year": 2014,
    "test_name": "E000012060A14F1",
    "code": 1206,
    "subject": "Psychology",
    "id": "E000012060A14F1-178"
}
\end{lstlisting}
\begin{center}
Listing 1: Example of JSON object (question) in the UNED-ACCESS 2024 dataset.
\label{ejemplo_json}
\end{center}
\end{minipage}
\end{center}


\section{Experimental Setup}
\label{sec:appendix-setup}

\subsection{Model Access}
\label{appendix-model access}
The proprietary models GPT-4-Turbo, GPT-4o, and GPT-3.5-Turbo were accessed via the OpenAI API, while Claude-3-Opus was available via Anthropic's API. Open-source models were available from HuggingFace (Llama-2-7B\footnote{\url{https://huggingface.co/google/gemma-7b-it}}, Llama-3-8B\footnote{\url{https://huggingface.co/meta-llama/Meta-Llama-3-8B-Instruct}}, Mistral-7B\footnote{\url{https://huggingface.co/mistralai/Mixtral-8x7B-Instruct-v0.1}}, Gemma-7B\footnote{\url{https://huggingface.co/google/gemma-7b-it}} and Leniachat-Gemma-2B\footnote{\url{https://huggingface.co/LenguajeNaturalAI/leniachat-gemma-2b-v0}}) or deployed via the Ollama library (Llama-3-70B\footnote{\url{https://ollama.com/library/llama3:70b-instruct}}, Mixtral-8x7B\footnote{\url{https://ollama.com/library/mixtral:instruct}} and Gemma-2-27B\footnote{\url{https://ollama.com/library/gemma2:27b-instruct-fp16}}).

\subsection{Instruction Format Templates}
\label{sec:appendice instruct}

Instruction following models often require instruction templates that format system, query and assistant messages with specific tags, available in the official documentation of each model. We have used this templates only when deploying models via Huggingface (all of them except proprietary models and the big ones deployed with Ollama, since ollama already includes these templates).

\begin{center}
\begin{lstlisting}
def _instruction_format(self, sys_message: str, query: str, assistant: str):
    if 'gemma' in self.model_id:
        return f'<start_of_turn>user
                {sys_message}
        
                {query}<end_of_turn>
                <start_of_turn>model
                {assistant}'
        
    elif 'mistral' in self.model_id or 'mixtral' in self.model_id:
        return f'<s> [INST] {sys_message} [/INST]\nUser: {query}\nAssistant: {assistant}'
    elif 'llama-2' in self.model_id:
        return f'<s>[INST] <<SYS>>
                {sys_message}
                <</SYS>>
                \n{query}</s>[/INST]
                \n<s>{assistant}</s>'
    elif 'llama-3' in self.model_id:
        return f'<|begin_of_text|><|start_header_id|>system<|end_header_id|>
                {sys_message}|<|eot_id|><|start_header_id|>user<|end_header_id|>
                {query}|<|eot_id|><|start_header_id|>assistant<|end_header_id|>
                {assistant}'
\end{lstlisting}
Listing 2: Python function showing instruction format templates for models that require so.
\label{templates}
\end{center}

\section{Results}

\subsection{Results on UNED-ACCESS 2024 by Subject}
\label{sec:appendix-results}

Table~\ref{base_asig} provides detailed results of the experiments, per language, model and discipline.

\subsection{Results on MMLU-10 by Subject}
\label{apendice_mmlu10}
Table~\ref{t:MMLU-subjects} provides the evaluation results on the  MMLU-10 dataset per subject.

\begin{table*}[p]
\centering
\resizebox{\linewidth}{!}{%
\renewcommand{\arraystretch}{0.8}
\begin{tabular}{lcccccccccccc}

 & \textbf{\scriptsize\begin{sideways}BAM\end{sideways}} & \textbf{\scriptsize\begin{sideways}Biology\end{sideways}} & \textbf{\scriptsize\begin{sideways}Biochemistry\end{sideways}} & \textbf{\scriptsize\begin{sideways}Economics\end{sideways}} & \textbf{\scriptsize\begin{sideways}F. of Computing\end{sideways}} & \textbf{\scriptsize\begin{sideways}Spanish Language\end{sideways}} & \textbf{\scriptsize\begin{sideways}Literature\end{sideways}} & \textbf{\scriptsize\begin{sideways}Mathematics\end{sideways}} & \textbf{\scriptsize\begin{sideways}Math Applied to SS\end{sideways}} & \textbf{\scriptsize\begin{sideways}Advanced Math  
\end{sideways}} & \textbf{\scriptsize\begin{sideways}Psychology\end{sideways}} & \textbf{\small Average}\\
\hline
SPANISH  &&&&&&&&&&&&\\
\hline
Claude-3-Opus & {\bf 0.86} & 0.96 & {\bf 1.00} & 0.90 & {\bf 0.98} & 0.67 & {\bf 0.84} & {\bf 0.69} & {\bf 0.63} & 0.44 & 0.89 & 0.81 \\

GPT-4-Turbo & 0.78 & 0.96 & {\bf 1.00} & {\bf 0.95} & 0.96 & 0.69 & 0.72 & 0.55 & 0.57 & {\bf 0.50} & 0.88 & 0.78 \\
GPT-4o & 0.84 & {\bf 0.97} & {\bf 1.00} & 0.90 & 0.96 & {\bf 0.74} & 0.81 & 0.51 & 0.38 & {\bf 0.50} & {\bf 0.91} & 0.77 \\

Llama-3-70B-Instruct & 0.83 & 0.89 & 0.95 & 0.79 & 0.94 & 0.39 & 0.66 & 0.38 & 0.46 & 0.25 & 0.82 & 0.67 \\
Gemma-2-27B-Instruct & 0.76 & 0.92 & {\bf 1.00} & 0.79 & 0.94 & 0.50 & 0.53 & 0.34 & 0.33 & 0.38 & 0.80 & 0.66 \\

Mixtral-8x7B-Instruct & 0.72 & 0.84 & 0.87 & 0.58 & 0.87 & 0.32 & 0.52 & 0.32 & 0.23 & 0.25 & 0.78 & 0.57 \\

GPT-3.5-Turbo & 0.64 & 0.80 & 0.90 & 0.53 & 0.87 & 0.32 & 0.44 & 0.20 & 0.20 & 0.38 & 0.74 & 0.55 \\

Llama-3-8B-Instruct & 0.57 & 0.71 & 0.82 & 0.56 & 0.79 & 0.26 & 0.37 & 0.22 & 0.30 & 0.25 & 0.67 & 0.50 \\

Mistral-7B-Instruct & 0.52 & 0.67 & 0.72 & 0.42 & 0.77 & 0.25 & 0.40 & 0.05 & 0.27 & 0.06 & 0.62 & 0.43 \\

Gemma-7B-It & 0.40 & 0.63 & 0.77 & 0.32 & 0.66 & 0.12 & 0.36 & 0.12 & 0.14 & 0.06 & 0.58 & 0.38 \\

Llama-2-7B-Chat & 0.29 & 0.41 & 0.31 & 0.19 & 0.56 & 0.12 & 0.34 & 0.14 & 0.12 & -0.12 & 0.44 & 0.25 \\

Leniachat-Gemma-2B & 0.19 & 0.21 & 0.03 & 0.06 & 0.15 & 0.05 & 0.22 & -0.03 & 0.20 & -0.12 & 0.24 & 0.11 \\
\hline
ENGLISH &&&&&&&&&&&&\\
\hline
Claude-3-Opus & \textbf{0.81} & 0.95 & \textbf{1.00} & 0.90 & \textbf{0.96} & 0.67 & 0.78 & \textbf{0.63} & \textbf{0.66} & 0.44 & 0.84 & 0.79 \\

GPT-4o & 0.79 & 0.96 & \textbf{1.00} & \textbf{0.92} & \textbf{0.96} & \textbf{0.70} & \textbf{0.82} & 0.55 & 0.52 & \textbf{0.50} & \textbf{0.86} & 0.78 \\

GPT-4-Turbo & 0.78 & \textbf{0.97} & \textbf{1.00} & \textbf{0.92} & 0.94 & 0.67 & 0.74 & 0.55 & 0.51 & \textbf{0.50} & 0.83 & 0.76 \\

Llama-3-70B-Instruct & 0.74 & 0.90 & \textbf{1.00} & 0.82 & 0.92 & 0.33 & 0.60 & 0.34 & 0.44 & 0.19 & 0.82 & 0.65 \\
Gemma-2-27B-Instruct & 0.72 & 0.94 & \textbf{1.00} & 0.79 & 0.87 & 0.50 & 0.55 & 0.28 & 0.33 & 0.25 & 0.81 & 0.64 \\
GPT-3.5-Turbo & 0.67 & 0.84 & 0.95 & 0.61 & 0.89 & 0.36 & 0.56 & 0.28 & 0.17 & 0.50 & 0.73 & 0.60 \\

Mixtral-8x7B-Instruct & 0.71 & 0.81 & 0.92 & 0.61 & 0.87 & 0.32 & 0.52 & 0.22 & 0.33 & 0.13 & 0.73 & 0.56 \\

Llama-3-8B-Instruct & 0.52 & 0.77 & 0.90 & 0.61 & 0.79 & 0.38 & 0.43 & 0.20 & 0.28 & 0.13 & 0.67 & 0.51 \\

Mistral-7B-Instruct & 0.57 & 0.71 & 0.82 & 0.63 & 0.77 & 0.23 & 0.36 & 0.05 & 0.23 & -0.00 & 0.65 & 0.46 \\

Gemma-7B-It & 0.41 & 0.67 & 0.85 & 0.56 & 0.75 & 0.18 & 0.22 & 0.20 & 0.14 & -0.06 & 0.61 & 0.41 \\

Llama-2-7B-Chat & 0.43 & 0.62 & 0.39 & 0.27 & 0.62 & 0.15 & 0.30 & 0.12 & 0.15 & -0.00 & 0.48 & 0.32 \\

Leniachat-Gemma-2B & 0.29 & 0.32 & 0.24 & 0.03 & 0.13 & 0.05 & 0.22 & 0.08 & 0.07 & -0.06 & 0.27 & 0.15 \\
\hline
\end{tabular}
}
\caption{Cohen's Kappa results on UNED-ACCESS 2024 by model and subject in Spanish and English, sorted by average. In bold, the best model(s) per subject per language.}
\label{base_asig}
\end{table*}

\begin{table*}[p]
\centering
\resizebox{\linewidth}{!}{%
\renewcommand{\arraystretch}{0.8}
\begin{tabular}{lccccccccccc}
 & \begin{turn}{90}\textbf{\scriptsize High School Biology}\end{turn} & \begin{turn}{90}\textbf{\scriptsize High School Chemistry}\end{turn} &  \begin{turn}{90}\textbf{\scriptsize High School Microeconomics}\end{turn} & \begin{turn}{90}\textbf{\scriptsize High School Macroeconomics}\end{turn} & \begin{turn}{90}\textbf{\scriptsize High School Computer Science}\end{turn} & \begin{turn}{90}\textbf{\scriptsize High School World History}\end{turn} &  \begin{turn}{90}\textbf{\scriptsize Elementary Mathematics}\end{turn} &  \begin{turn}{90}\textbf{\scriptsize High School Mathematics}\end{turn} & \begin{turn}{90}\textbf{\scriptsize High School Physics}\end{turn} &  \begin{turn}{90}\textbf{\scriptsize High School Psychology}\end{turn} & \textbf{\small Average} \\
\midrule
GPT-4o & {\bf 0.94} & {\bf 0.72} & {\bf 0.95} & {\bf 0.88} & {\bf 0.89} & {\bf 0.93} & 0.64 & 0.33 & {\bf 0.62} & {\bf 0.94} & {\bf 0.79} \\
Claude-3-Opus & 0.91 & 0.67 & 0.89 & 0.81 & {\bf 0.89} & 0.91 & {\bf 0.86} & {\bf 0.40} & 0.56 & 0.93 & 0.78 \\
GPT-4-Turbo & 0.90 & 0.65 & 0.92 & 0.83 & 0.88 & 0.89 & 0.62 & 0.36 & 0.57 & 0.93 & 0.75 \\
Gemma2-27B-Instruct & 0.91 & 0.62 & 0.84 & 0.79 & 0.77 & 0.89 & 0.44 & 0.29 & 0.43 & 0.91 & 0.69 \\
Llama3-70B-Instruct  & 0.83 & 0.47 & 0.84 & 0.73 & 0.76 & 0.82 & 0.63 & 0.29 & 0.46 & 0.90 & 0.67 \\
Mixtral-7B-Instruct & 0.75 & 0.42 & 0.64 & 0.57 & 0.59 & 0.82 & 0.30 & 0.17 & 0.30 & 0.81 & 0.54 \\
GPT-3.5-Turbo & 0.72 & 0.38 & 0.63 & 0.55 & 0.64 & 0.74 & 0.35 & 0.15 & 0.20 & 0.81 & 0.52 \\
Llama-3-8B-Instruct & 0.71 & 0.36 & 0.64 & 0.53 & 0.57 & 0.75 & 0.25 & 0.04 & 0.23 & 0.79 & 0.49 \\
Mistral-7B-Instruct & 0.56 & 0.24 & 0.50 & 0.37 & 0.45 & 0.67 & 0.17 & 0.11 & 0.11 & 0.71 & 0.39 \\
Gemma-7B-It & 0.47 & 0.15 & 0.36 & 0.33 & 0.31 & 0.50 & -0.12 & -0.25 & -0.05 & 0.63 & 0.23 \\
Llama-2-7B-Chat & 0.22 & 0.09 & 0.10 & 0.10 & 0.13 & 0.30 & 0.06 & -0.04 & 0.01 & 0.38 & 0.13 \\
Leniachat-Gemma-2B & 0.13 & 0.04 & 0.08 & 0.09 & 0.07 & 0.08 & 0.03 & -0.04 & -0.04 & 0.21 & 0.06 \\
\bottomrule
\end{tabular}
}
\caption{Cohen's Kappa performance of models per subject on MMLU-10.\label{t:MMLU-subjects}}
\end{table*}

\end{document}